\documentclass{article}
\usepackage{arxiv}
\usepackage{mathtools}
\usepackage{graphicx}
\usepackage{hyperref}
\usepackage[binary-units]{siunitx}
\usepackage{booktabs}
\usepackage{xspace}
\usepackage{csquotes}
\usepackage[frozencache,cachedir=.]{minted}
\usepackage{algorithm}
\usepackage{pgfplotstable}
\usetikzlibrary{shapes}
\usetikzlibrary{patterns}
\usepackage[outline]{contour}
\usepackage{pgfplots}
\usepackage{colortbl}
\usepackage{notation}

\DeclareNum{\numinstances}{n}
\DeclareNum{\numlabels}{m}
\DeclareNum{\nummeta}{b}
\DeclareNum{\numhidden}{e}
\DeclareNum{\batchsize}{b}
\DeclareNum{\numhashings}{r}
\DeclareFun{\hashfun}{h}
\DeclareFun{\classifier}{f}
\DeclareFun{\binloss}{\ell}
\DeclareFun{\featureextract}{\psi}
\DeclareVecFun{\machcls}{\phi}

\DeclareVecFun{\concatmachcls}{\varphi}

\DeclareSet{\instancespace}{X}
\DeclareAbstract{\sinstance}{x}
\DeclareRandVar{\rinstance}{x}

\DeclareVec{\labelvec}{y}
\DeclareRandVar{\rlabelvec}{y}
\DeclareVec{\metalabelvec}{g}

\DeclareModification{\predictvec}{withhat}{\labelvec}

\DeclareNum{\hashindex}{s}
\DeclareNum{\labelindex}{j}
\DeclareNum{\metaindex}{i}
\DeclareMat{\indicatormat}{C}
\DeclareAbstract{\indicatormatcmp}{c}
\DeclareMat{\weightmatrix}{W}
\DeclareVec{\weightvector}{w}
\DeclareMat{\weightmatrixd}{W}

\NewDocumentCommand{\topk}{o}{
    \IfNoValueTF{#1}{\operatorname{top}}{\operatorname{top}_#1}
}
\NewDocumentCommand{\patk}{o}{
    \operatorname{P}\IfNoValueTF{#1}{}{\!@{#1}}
}

\newcommand{\tensorflow}{\texttt{tensorflow}\xspace}
\DeclareDataset{\slice}{Slice}
\DeclareDataset{\cascade}{Cascade}
\DeclareDataset{\wikiten}{Wiki10-31k}
\DeclareDataset{\wikitencascade}{Wiki10-31k-cascade}
\DeclareDataset{\wikifk}{Wikipedia-500k}
\DeclareDataset{\amazonsix}{Amazon-670k}
\DeclareDataset{\wikifkcascade}{Wikipedia-500k-Cascade}
\DeclareDataset{\wikifkslice}{Wikipedia-500k-Slice}

\title{
Towards Memory-Efficient Training for Extremely Large Output Spaces -- Learning with 500k Labels on a Single Commodity GPU
}

\renewcommand{\shorttitle}{Training for Deep Extreme Classifiers on Single Commodity GPU}
\author{
    Erik Schultheis \\
    Aalto University\\
    Helsinki, Finland \\
	\texttt{erik.schultheis@aalto.fi} \\
	\And
    Rohit Babbar \\
    University of Bath / Aalto University\\
    Bath, UK / Helsinki, Finland \\
	\texttt{rb2608@bath.ac.uk} 
}

\begin{document}
\date{}

\maketitle              
\begin{abstract}
In classification problems with large output spaces (up to millions of labels),
the last layer can require an enormous amount of memory. Using sparse
connectivity would drastically reduce the memory requirements, but as we show
below, it can result in much diminished predictive performance of the model.
Fortunately, we found that this can be mitigated by introducing a penultimate
layer of intermediate size. We further demonstrate that one can constrain the
connectivity of the sparse layer to be uniform, in the sense that each output
neuron will have the exact same number of incoming connections. This allows for
efficient implementations of sparse matrix multiplication and connection
redistribution on GPU hardware. Via a custom CUDA implementation, we show that the
proposed approach can scale to datasets with 670,000 labels on a single commodity GPU with only 4GB memory.

\end{abstract}
\section{Introduction}

In this paper, we present findings towards employing sparse connectivity in
order to reduce the memory consumption of the classification layer for problems
with extremely large output spaces (XMC). Such problems arise in, e.g., tagging
of text documents~\cite{Dekel_Shamir_2010}, next-word
predictions~\cite{mikolov2013efficient}, and different kinds of recommendation
tasks~\cite{Beygelzimer_et_al_2009b,Weston_et_al_2013,Agrawal_et_al_2013,Prabhu_Varma_2014,medini_extreme_2019}.
In order to ensure computational tractability of these tasks, which can have up
to several millions of labels, one typically builds a \emph{hierarchical label
tree}~%
\cite{prabhu_parabel_2018,you_attentionxml_2019,jiang_lightxml_2021,wydmuch2018no,khandagale2020bonsai},
only exploring branches that are likely to contain relevant labels for the
current instance. Even though this is very effective at reducing the
computation (from linear in the number of labels to logarithmic), it does
not help in addressing the memory consumption, which is still linear in the
number of labels times the number of hidden units.

As an illustration consider the \textsc{Amazon-3M}~\cite{mcauley2015inferring}
dataset. If we were to map the inputs to a hidden representation of \num{1024}
units, the fully connected last layer for this dataset would need about
\num{2.9} billion parameters, corresponding to \SI{10.7}{\gibi
\byte}\footnote{Assuming 32-bit floating point numbers}. Given that modern deep
learning optimizers such as \textsc{Adam}~\cite{kingma2014adam} need to keep
track of the value, gradient, and first and second moment, this leads to an
overall peak memory consumption of over \SI{40}{\gibi \byte}, making it nigh
impossible to train such models on commodity hardware.

Therefore, we want to investigate possibilities for memory efficient \emph{sparse training} of
this huge last layer. There are two pre-existing approaches that serve as an
indication that this is an idea that could be successful: First, for
\textsc{DiSMEC}, a \emph{linear model} applied to tf-idf representations of
input text, it is known that the resulting layer can be sparsified \emph{after
training} to contain \emph{less than 1\%} non-zeros~\cite{babbar_dismec_2017}.
In a linear model, the different classifiers for each label can be trained
\emph{independently}. As a result, only the full weights of the label that is
currently trained needs to be kept in memory, and can be pruned as soon as the
training for that label has finished. For non-linear models, the
\textsc{Mach}~\cite{medini_extreme_2019} algorithm can be interpreted as a
special case of training with \emph{static}, random sparsity. It works by
hashing the labels into different buckets, and performing training and
predictions only on the level of buckets. If enough independent hashes are used,
this method allows to solve the original problem in the large output space.
However, in practice, the results presented for \textsc{Mach} are not as good as
for competing methods.

The contributions of this paper are as follows: We show that na{\"i}vely
applying a dynamic sparse training algorithm to the last layer of an XMC problem
results in strongly reduced predictive performance. Inspired by \textsc{Mach},
we then propose to alleviate this problem by inserting a penultimate layer that
is larger than the hidden representation of the inputs, but still much smaller
than the size of the label space. Such an increased layer size drastically
improves the chances of dynamic sparse training finding a good subnetwork, and
enables us to get results only slightly worse than training with a dense last
layer. We demonstrate this on several large-scale datasets. To ensure memory
efficient and quick computations, we propose to restrict the sparsity structure
to \emph{uniform sparsity}, such that each unit in the output layer receives exactly the same
number of inputs. This has several important consequences : (i) it makes it
impossible for the training to focus most non-zero weights on a few, prominent
head labels, and instead ensures a more even distribution of the
representational capacity, (ii) compared to
\href{https://www.intel.com/content/www/us/en/docs/onemkl/developer-reference-c/2023-0/sparse-blas-coordinate-matrix-storage-format.html}{coordinate-format}
this requires only half the memory to store the indices, and compared to
\href{https://www.intel.com/content/www/us/en/docs/onemkl/developer-reference-c/2023-0/sparse-blas-csr-matrix-storage-format.html}{compressed
row sparse} matrices the data layout is simpler, making it easier to implement
the corresponding operations on a GPU, and 
 (iii)  it also
means that changing the sparsity structure (redistribution of connections) can
be implemented as a very cheap operation.

\section{Setup and Background}

We consider classification problems that map an input instance $\sinstance
\in \instancespace$ to a subset of a label set with $\numlabels$ labels,
represented as a binary vector $\labelvec \in \set{0, 1}^{\numlabels}$. More
precisely, we assume that $(\rinstance, \labelvec) \sim \mathds{P}$ are jointly
distributed according to some probability measure.  If almost surely
$\|\labelvec\|_1 = 1$, it is a multiclass setup, otherwise a multilabel setup.
We want to find a classifier
$\defmap{\classifier}{\instancespace}{\set{0,1}^{\numlabels}}$ so that predicted
and actual labels are close. Usually, $\classifier$ can be decomposed into two
operations: First, the inputs are \emph{embedded} into a fixed-size vector space
using a function $\defmap{\featureextract}{\instancespace}{\reals^{\numhidden}}$ (e.g. a
linear projection, multilayer perceptron, or transformer-based text model), and
then a decoding $\weightmatrix \in \reals^{\numhidden \times \numlabels}$ is applied to
extract scores for each label. The actual prediction is then generated by
selecting the $k$ highest scoring labels as positive, $\predictvec =
\topk[k](\weightmatrix^{\mathsf{T}} \featureextract(\sinstance))$.
Consequently, performance is typically measured in terms of
\emph{precision-at-$k$}, defined as the fraction of correct predictions
\begin{equation}
    \patk[k](\labelvec, \predictvec) = k^{-1} \sum_{j=1}^{\numlabels} \labelvec[j] \predictvec[j] \qquad \text{for $\|\predictvec\|_1 = k$}\,.
\end{equation}

In order to find the optimal $\weightmatrix$ that maximizes $\patk[k]$, one
often performs a \emph{One-vs-All (OvA)}
reduction\cite{babbar_dismec_2017,proxml,menon_multilabel_2019}: A binary classification
loss $\binloss$ is applied to each label separately. As this involves predicting
the scores $\weightmatrix^{\mathsf{T}} \featureextract(\sinstance)$ for each
label, many methods select a subset $\mathcal{N} \subset \intrange{\numlabels}$
of \emph{hard
negatives}\cite{jain2019slice,Chang_et_al_2020,kharbanda2022cascadexml,reddi2019stochastic,jiang_lightxml_2021},
to approximate the sum as
\begin{multline}
    l(\labelvec, \sinstance)
    = \sum_{j=1}^{\numlabels} \binloss(\labelvec[j], \weightvector_j^{\mathsf{T}} \featureextract(\sinstance)) = \sum_{j: \labelvec[j] = 1} \binloss(1, \weightvector_j^{\mathsf{T}} \featureextract(\sinstance)) + \sum_{j: \labelvec[j] = 0} \binloss(0, \weightvector_j^{\mathsf{T}} \featureextract(\sinstance)) \\
    \approx \sum_{j: \labelvec[j] = 1} \binloss(1, \predictvec[j]) + \sum_{j \in \mathcal{N}} \binloss(0, \predictvec[j]) \,.
\end{multline}
This is very effective in reducing the required computations, and could also be
beneficial because it effectively changes the distribution of labels seen by the
classifier \cite{disentangling-rawat21a}, but it does not decrease the enormous
amount of memory required to store the weight matrix $\weightmatrix$.

There are several established approaches to handle this problem: The most straightforward
method is to place a \emph{bottleneck} layer just before the final classification
layer, so that the dimension of the embedding that $\weightmatrix$ operates
on is comparatively low. For example, LightXML\cite{jiang_lightxml_2021} project
the \num{3280}-dimensional representation used for determining hard negatives
down to only \num{300} units for the extreme-level classification. This approach
is limited in its effectiveness, as too small sizes start to severely affect
the classification quality. 
A second strategy is to \emph{prune} the matrix $\weightmatrix$ after training,
turning it into a very sparse matrix. This can reduce the model size to only a
tiny fraction of the dense equivalent, without negatively affecting its
predictive power, but this does not solve the problem of memory consumption
during the training itself. The only exception are linear models, where the
weight vectors $\weightvector_j$ for different labels can be trained
independently, and be sparsified immediately after training, so that the full
matrix never has to materialize\cite{babbar_dismec_2017,proxml}. Additionally,
it is possible to exploit the relation between primal and dual of linear
problems to achieve sparse training for max margin classifiers with appropriate
loss functions\cite{yen2016pd}. Finally, MACH\cite{medini_extreme_2019} has shown
that it is possible to train an extreme classifier on the level of meta-labels,
obviating the need for the large weight matrix $\weightmatrix$ altogether.
As shown in the next section, this corresponds, implicitly, to a multiplication by a sparse, but \textit{fixed},
binary matrix which therefore limits the expressiveness of the model.

Thus, existing sparse training methods for XMC either use post-training
sparsification, or a fixed sparsity structure. Here, we want to apply the
\emph{sparse evolutionary training (\textsc{Set})}
algorithm\cite{mocanu_scalable_2018} to the classification layer, so that we
have sparse training with dynamic sparsity structure. The \textsc{Set} algorithm
follows a general prune-redistribute-regrowth cycle, which means that
periodically, a subset of existing non-zero weights is selected to be removed
(\emph{pruned}), and new structural non-zeros will be inserted
(\emph{redistributed}). After that, the training of the sparse layer proceeds
just as in any other gradient-descent based optimization, i.e., the structural
non-zeros are updated according to their mini-batch gradient (\emph{regrown}),
and the structural zeros are left unchanged, until the next cycle starts.

This general algorithmic structure can be implemented in various ways, depending
on how the pruned weights are selected, and how it is determined where they
should be re-distributed. The \textsc{Set} algorithm uses very simple
heuristics: The set of least important connections is determined by sorting
according to the absolute value of their weight, and removing the fraction
$\alpha$ of connections with lowest weight. The same number of new connections
is inserted after pruning, by choosing uniformly randomly from the structural
zeros. 

While there exist other elaborate schemes, they are generally more complex to implement
and will require additional memory. For example, \cite{deepr} chooses its pruning
based on weights switching their sign, which means that it needs to store the previous
signs of all structural non-zeros. To determine useful locations for inserting 
the redistributed connections, \cite{dettmers2019sparse} uses a momentum term,
which means that this requires the same amount of memory as the weights for
the original dense layer, and thus is infeasible in our setting. This also
excludes any strategy that requires, even if only intermittently, a full, dense 
gradient to be computed, such as \cite{evci2020rigging}.

A na\"ive application of \textsc{Set} to the last layer leads to unsatisfactory
results, and an implementation using just the available tools in
\texttt{tensorflow} turns out to be suboptimal in terms of speed and memory
consumption. Thus, we present in the next section some modifications to the
architecture and training algorithm, as well as some insights into an efficient
implementation, that alleviate these shortcomings.

\section{Method}

\label{sec:method}

In principle, implementing a sparse layer in \tensorflow\footnote{At the
time of this writing, \texttt{PyTorch} still considers its sparse tensor support
to be in beta.} is straightforward: Just replace the dense-dense matrix
multiplication with a corresponding sparse-dense operation that is supplied by
the framework, and replace the dense weight matrix with a \texttt{SparseTensor}
object.

There are four problems with this approach: First, it wastes memory due to
{\tensorflow}s requirement that all indices be given as 64-bit integers. Second,
completely unstructured sparsity makes efficient implementations challenging.
Third, the \tensorflow operations cannot exploit the sparsity in the gradient
signal that arises naturally when training with hinge-like losses. Finally,
replacing the dense layer with a highly sparse layer results in underfitting. We
will address these problems below.

\subsection{Efficient 32-bit indexing}
In \tensorflow, sparse tensors are represented in \emph{coordinate (COO)} format,
which means that each structural nonzero in a sparse matrix is described by 
three numbers. Two 64-bit integers define the row and column of the structural
nonzero, and a 32-bit floating point number its value. This means that a single
sparse weight requires as much memory as five weights in the dense matrix.

Even for extreme-scale classification, however, 32-bit integers would be more
than sufficient as column and row indices of $\weightmatrix$. A maximum
representable value of around 4 billion is still an order of magnitude larger
than even very large scale proprietary problems\cite{medini_extreme_2019} with
100s of millions of labels, and three orders of magnitude larger than publicly
available benchmark datasets.

\begin{figure}
\centering

\begin{tikzpicture}[circular/.style={circle,fill, minimum size=0.45cm, inner sep=0, text=black}]

\node at (1.5, 3.4) {\small a) COO};
\node at (5.5, 3.4) {\small b) CSC};
\node at (9.5, 3.4) {\small c) Uniform};

\newcommand{\wgtmtx}{
\node at (1.5, 2.8) {\scriptsize label};
\node[rotate=90] at (-0.3, 1.25) {\scriptsize feature};
\node[red!33!white, circular] at (0.5, 2) {$\scriptscriptstyle 0.7$};
\node[orange!33!white, circular] at (0.5, 1) {$\scriptscriptstyle 1.2$};
\node[blue!33!white, circular] at (1.0, 1.5) {$\scriptscriptstyle -1$};
\node[green!33!white, circular] at (2.0, 2.0) {$\scriptscriptstyle 2.2$};
\node[olive!33!white, circular] at (2.0, 0.5) {$\scriptscriptstyle 0.0$};
\node[violet!33!white, circular] at (2.5, 0.5) {$\scriptscriptstyle 0.3$};
\foreach \i in {0,...,4} {
    \node at (0.5 + 0.5*\i, 2.45) {\scriptsize$\i$};
}
\foreach \i in {0,...,3} {
    \node at (0.05, 2.0 - 0.5*\i) {\scriptsize$\i$};
}
\begin{scope}[shift={(0.25,0.25)}]
\draw[step=0.5cm,black,very thin] (0,0) grid (2.5,2.0);
\draw[thick] (0, 0) -- (2.5, 0);
\draw[thick] (2.5, 0) -- (2.5, 2);
\draw[thick] (2.5, 2) -- (0, 2);
\draw[thick] (0, 0) -- (0, 2);
\end{scope}
}

\wgtmtx

\begin{scope}[shift={(0,-1.6)}]
\begin{scope}[shift={(0.25,0.25)}]
\draw[step=0.5cm,black,very thin] (0,0) grid (3.0,1.0);
\draw[thick] (0, 0) -- (3, 0);
\draw[thick] (3, 0) -- (3, 1);
\draw[thick] (3, 1) -- (0, 1);
\draw[thick] (0, 0) -- (0, 1);
\end{scope}
\node at (1.8, 1.5) {\scriptsize indices};
\node at (-0.25, 1.0) {\scriptsize col};
\node at (-0.25, 0.5) {\scriptsize row};
\node[red!33!white, circular] at (0.5, 1.0) {$0$};
\node[red!33!white, circular] at (0.5, 0.5) {$0$};
\node[orange!33!white, circular] at (1, 1.0) {$0$};
\node[orange!33!white, circular] at (1, 0.5) {$2$};
\node[blue!33!white, circular] at (1.5, 1.0) {$1$};
\node[blue!33!white, circular] at (1.5, 0.5) {$1$};
\node[green!33!white, circular] at (2.0, 1.0) {$3$};
\node[green!33!white, circular] at (2.0, 0.5) {$0$};
\node[olive!33!white, circular] at (2.5, 1.0) {$3$};
\node[olive!33!white, circular] at (2.5, 0.5) {$3$};
\node[violet!33!white, circular] at (3.0, 1.0) {$4$};
\node[violet!33!white, circular] at (3.0, 0.5) {$3$};
\end{scope}

\begin{scope}[shift={(0,-2.7)}]
\begin{scope}[shift={(0.25,0.25)}]
\draw[step=0.5cm,black,very thin] (0,0) grid (3.0,0.5);
\draw[thick] (0, 0) -- (3, 0);
\draw[thick] (3, 0) -- (3, 0.5);
\draw[thick] (3, 0.5) -- (0, 0.5);
\draw[thick] (0, 0) -- (0, 0.5);
\end{scope}

\begin{scope}[shift={(0.5,0.5)}]
\node at (1.3, 0.5) {\scriptsize weights};
\node[red!33!white, circular] at (0.0, 0.0) {$\scriptscriptstyle 0.7$};
\node[orange!33!white, circular] at (0.5, 0.0) {$\scriptscriptstyle 1.2$};
\node[blue!33!white, circular] at (1.0, 0.0) {$\scriptscriptstyle -1$};
\node[green!33!white, circular] at (1.5, 0.0) {$\scriptscriptstyle 2.2$};
\node[olive!33!white, circular] at (2.0, 0.0) {$\scriptscriptstyle 0.0$};
\node[violet!33!white, circular] at (2.5, 0.0) {$\scriptscriptstyle 0.3$};
\end{scope}
\end{scope}

\begin{scope}[shift={(4,0.0)}]
\wgtmtx

\begin{scope}[shift={(0,-1.1)}]
\begin{scope}[shift={(0.25,0.25)}]
\draw[step=0.5cm,black,very thin] (0,0) grid (2.5,0.5);
\draw[thick] (0, 0) -- (2.5, 0);
\draw[thick] (2.5, 0) -- (2.5, 0.5);
\draw[thick] (2.5, 0.5) -- (0, 0.5);
\draw[thick] (0, 0) -- (0, 0.5);
\draw[step=0.5cm,black,very thin,shift={(0,0.2)}] (0,-1) grid (3.0,-0.5);
\draw[thick] (0, -0.3) -- (3, -0.3);
\draw[thick] (3, -0.3) -- (3, -0.8);
\draw[thick] (3, -0.8) -- (0, -0.8);
\draw[thick] (0, -0.3) -- (0, -0.8);

\draw[->] (0.25,0) -- (0.25,-0.3);
\draw[->] (0.75,0) -- (1.25,-0.3);
\draw[->] (1.25,0) -- (1.75,-0.3);
\draw[->] (1.75,0) -- (1.75,-0.3);
\draw[->] (2.25,0) -- (2.75,-0.3);
\end{scope}
\node at (1.4, 1.0) {\scriptsize rowptr};
\node at (0.5, 0.5) {$0$};
\node at (1.0, 0.5) {$2$};
\node at (1.5, 0.5) {$3$};
\node at (2.0, 0.5) {$3$};
\node at (2.5, 0.5) {$5$};

\begin{scope}[shift={(0.5,-0.3)}]
\node[red!33!white, circular] at (0, 0.0) {$0$};
\node[orange!33!white, circular] at (0.5, 0.0) {$2$};
\node[blue!33!white, circular] at (1, 0.0) {$1$};
\node[green!33!white, circular] at (1.5, 0.0) {$0$};
\node[olive!33!white, circular] at (2, 0.0) {$3$};
\node[violet!33!white, circular] at (2.5, 0.0) {$3$};
\end{scope}
\end{scope}

\begin{scope}[shift={(0,-2.7)}]
\begin{scope}[shift={(0.25,0.25)}]
\draw[step=0.5cm,black,very thin] (0,0) grid (3.0,0.5);
\draw[thick] (0, 0) -- (3, 0);
\draw[thick] (3, 0) -- (3, 0.5);
\draw[thick] (3, 0.5) -- (0, 0.5);
\draw[thick] (0, 0) -- (0, 0.5);
\end{scope}
\begin{scope}[shift={(0.5,0.5)}]
\node[red!33!white, circular] at (0.0, 0.0) {$\scriptscriptstyle 0.7$};
\node[orange!33!white, circular] at (0.5, 0.0) {$\scriptscriptstyle 1.2$};
\node[blue!33!white, circular] at (1.0, 0.0) {$\scriptscriptstyle -1$};
\node[green!33!white, circular] at (1.5, 0.0) {$\scriptscriptstyle 2.2$};
\node[olive!33!white, circular] at (2.0, 0.0) {$\scriptscriptstyle 0.0$};
\node[violet!33!white, circular] at (2.5, 0.0) {$\scriptscriptstyle 0.3$};
\end{scope}
\end{scope}
\end{scope}

\begin{scope}[shift={(8,0.0)}]
\node at (1.5, 2.8) {\scriptsize label};
\node[rotate=90] at (-0.3, 1.25) {\scriptsize feature};
\node[red!33!white, circular] at (0.5, 2) {$\scriptscriptstyle 0.7$};
\node[orange!33!white, circular] at (0.5, 1) {$\scriptscriptstyle 1.2$};
\node[blue!33!white, circular] at (1.0, 1.5) {$\scriptscriptstyle -1$};
\node[purple!33!white, circular] at (1.0, 0.5) {$\scriptscriptstyle 0.6$};
\node[pink!33!white, circular] at (1.5, 2.0) {$\scriptscriptstyle 2.4$};
\node[magenta!33!white, circular] at (1.5, 1.5) {$\scriptscriptstyle 3.1$};
\node[green!33!white, circular] at (2.0, 2.0) {$\scriptscriptstyle 2.2$};
\node[olive!33!white, circular] at (2.0, 0.5) {$\scriptscriptstyle 0.0$};
\node[yellow!33!white, circular] at (2.5, 1.0) {$\scriptscriptstyle 0.2$};
\node[violet!33!white, circular] at (2.5, 0.5) {$\scriptscriptstyle 0.3$};
\foreach \i in {0,...,4} {
    \node at (0.5 + 0.5*\i, 2.45) {\scriptsize$\i$};
}
\foreach \i in {0,...,3} {
    \node at (0.05, 2.0 - 0.5*\i) {\scriptsize$\i$};
}
\begin{scope}[shift={(0.25,0.25)}]
\draw[step=0.5cm,black,very thin] (0,0) grid (2.5,2.0);
\draw[thick] (0, 0) -- (2.5, 0);
\draw[thick] (2.5, 0) -- (2.5, 2);
\draw[thick] (2.5, 2) -- (0, 2);
\draw[thick] (0, 0) -- (0, 2);
\end{scope}

\begin{scope}[shift={(0,-1.6)}]
\begin{scope}[shift={(0.25,0.25)}]
\draw[step=0.5cm,black,very thin] (0,0) grid (2.5,1.0);
\draw[thick] (0, 0) -- (2.5, 0);
\draw[thick] (2.5, 0) -- (2.5, 1);
\draw[thick] (2.5, 1) -- (0, 1);
\draw[thick] (0, 0) -- (0, 1);
\end{scope}
\node at (1.5, 1.5) {\scriptsize indices};
\begin{scope}[shift={(0.5,0.5)}]
\node[red!33!white, circular] at (0, 0.5) {$0$};
\node[orange!33!white, circular] at (0, 0.0) {$2$};
\node[blue!33!white, circular] at (0.5, 0.5) {$1$};
\node[purple!33!white, circular] at (0.5, 0.0) {$3$};
\node[pink!33!white, circular] at (1.0, 0.5) {$0$};
\node[magenta!33!white, circular] at (1.0, 0.0) {$1$};
\node[green!33!white, circular] at (1.5, 0.5) {$0$};
\node[olive!33!white, circular] at (1.5, 0.0) {$3$};
\node[yellow!33!white, circular] at (2.0, 0.5) {$2$};
\node[violet!33!white, circular] at (2.0, 0.0) {$3$};
\end{scope}
\end{scope}

\begin{scope}[shift={(0,-3.2)}]
\begin{scope}[shift={(0.25,0.25)}]
\draw[step=0.5cm,black,very thin] (0,0) grid (2.5,1.0);
\draw[thick] (0, 0) -- (2.5, 0);
\draw[thick] (2.5, 0) -- (2.5, 1);
\draw[thick] (2.5, 1) -- (0, 1);
\draw[thick] (0, 0) -- (0, 1);
\end{scope}
\node at (1.5, 1.5) {\scriptsize weights};
\begin{scope}[shift={(0.5,0.5)}]
\node[red!33!white, circular] at (0, 0.5) {$\scriptstyle 0.7$};
\node[orange!33!white, circular] at (0, 0.0) {$\scriptstyle 1.2$};
\node[blue!33!white, circular] at (0.5, 0.5) {$\scriptstyle -1$};
\node[purple!33!white, circular] at (0.5, 0.0) {$\scriptstyle 0.6$};
\node[pink!33!white, circular] at (1.0, 0.5) {$\scriptstyle 2.4$};
\node[magenta!33!white, circular] at (1.0, 0.0) {$\scriptstyle 3.1$};
\node[green!33!white, circular] at (1.5, 0.5) {$\scriptstyle 2.2$};
\node[olive!33!white, circular] at (1.5, 0.0) {$\scriptstyle 0.0$};
\node[yellow!33!white, circular] at (2.0, 0.5) {$\scriptstyle 0.2$};
\node[violet!33!white, circular] at (2.0, 0.0) {$\scriptstyle 0.3$};
\end{scope}
\end{scope}

\end{scope}

\end{tikzpicture}
\caption{Schematic depiction of different sparse matrix formats. Note that in COO
format (Fig.~\ref{fig:sparse-formats}a), the \texttt{indices} array in \autoref{alg:uniform-sparse-forward} is of shape $2 \times \text{nnz}$.  
In uniform format (Fig.~\ref{fig:sparse-formats}c) it is $\text{nnz per column} \times \text{labels}$, and hence
only half as big, compared to the COO format, for the same number of nonzeros.\label{fig:sparse-formats}}
\end{figure}
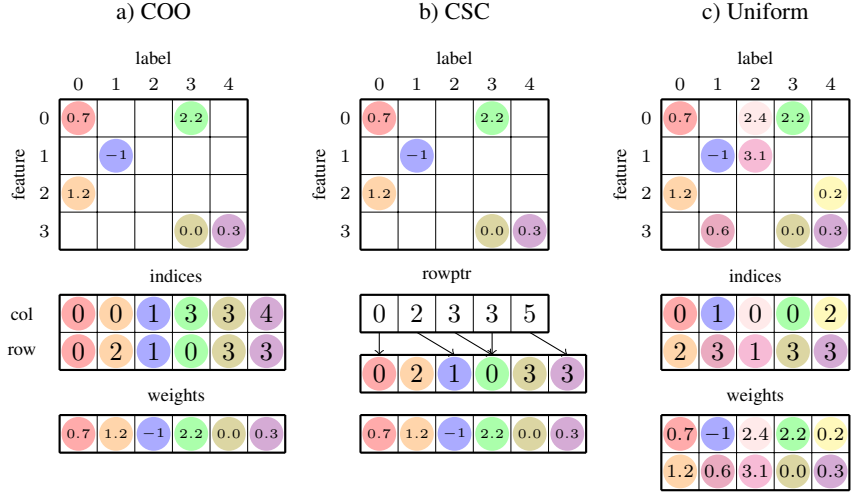

\subsection{Compressed indexing and equitable work distribution through uniform sparsity}
Even with 32-bit indices, a sparse weight still consumes three times as much
memory as a dense weight, when represented in coordinate format. This could be
much more efficient by switching to \emph{compressed sparse column (CSC)}
format, where only row indices are saved directly, and for each column only the
offset of its first index is stored. While this drastically reduces the amount
of memory needed to store the indices, it also increases the complexity of
involved computations. For example, in COO format, one can assign each thread on
the GPU with the same amount of structural non-zeros to handle during the matrix
multiplication, as getting the corresponding row and column indices is a simple
array lookup. In contrast, in CSC format, it is still trivial to assign one
column to each thread (i.e., each thread will compute one output), but that can
lead to a significant difference in the amount of work each thread has to do,
and thus lead to inefficient use of GPU resources. Furthermore, redistribution
becomes more involved, as inserting a new structural nonzero in an early column
means that all the weights and indices that come after have to be shifted.

This can be simplified if we stipulate that each column should have the exact
same amount of structural non-zeros, such that $\forall j:
\|{\weightvector}_j\|_0 = s$. Then, a single index array is sufficient, and the
starting offset of each column can be calculated simply by multiplying the
number of non-zeros per column with the column index, like in multidimensional
regular array indexing. Distributing a multiplication with a uniformly-sparse
matrix across many threads is also easy, as we can simply assign one column
(i.e., ${\weightvector}_j$) to each thread, knowing that they correspond to the
same amount of work. Finally, connection redistribution is much
easier, because the number of non-zeros stays constant for each column, and thus
changes in one column never require moving around the data of other columns. 
As we will show in Section~\ref{subsec:vary-arch}, the additional constraint on
the number of connections per output does not negatively influence the models
predictive performance.

Broadly, the implementation works as follows: The sparse weights are represented
by two matrices, $\texttt{indices} \in \naturals^{s \times \numlabels}$ and
$\texttt{weights} \in \reals^{s \times \numlabels}$. The input is given as 
a matrix $\texttt{features} \in \reals^{\batchsize \times \numhidden}$, where 
$\batchsize$ denotes the batch size, and the output will be a matrix
$\texttt{output} \in \reals^{\batchsize \times \numlabels}$. 
CUDA threads are generated on a two dimensional
grid, with one thread for each output that is to be calculated. Thus,
threads will be indexed by pairs, each of them consisting of $\texttt{instance} \in \intrange{\batchsize}$
and $\texttt{label} \in \intrange{\numlabels}$. Concretely, every thread performs 
the calculations given, schematically, in \autoref{alg:uniform-sparse-forward}.
\begin{algorithm}
\begin{minted}{python}
value = 0;
for weight_idx in range(s):
    source = indices[weight_idx, label]
    feature = features[instance, source]
    value += feature * weights[weight_idx, label]
output[instance, label] = value
\end{minted}
\caption{Calculation of the score for a single label \texttt{label} and \texttt{instance} for uniform sparsity (see Fig.~\ref{fig:sparse-formats}c)
with \texttt{s} non-zeros for each label.\label{alg:uniform-sparse-forward}}
\end{algorithm}

\subsection{Speeding up backward pass through implicit negative mining}
Our experiments with a sparse last layer showed that the largest fraction
of time was spent in the backward pass. This is not surprising, as the backward
pass actually requires two sparse matrix multiplications, to calculate the
gradient with respect to the inputs, and to calculate the gradient with respect
to the weights. 

Certain margin-based losses can induce high amounts of sparsity in the gradient
of XMC problems, which can be exploited to ensure considerable
speed-up\cite{schultheis2022speeding,yen2016pd}. In the given enormous label
space, each instance will have only a tiny subset of labels which are relevant
to it, and many for which the decision that they are not relevant is
\textquote{easy}. Thus, if the loss function gives zero penalty for these easy
classifications (e.g., if the margin is large enough in hinge-like losses), then
the error term to be back-propagated will be highly sparse. For the loss
function that is mainly used in this paper, the squared-hinge loss $\ell(y,
\hat{y}) = \max(0, 1 - y \hat{y})^2$, the gradient is $\partial \ell /
\partial \hat{y} = -2y \max(0, 1 - y \hat{y})$, and thus exactly zero whenever
$y \hat{y} \geq 1$.

Thus, in the backward kernel, it becomes beneficial to explicitly check whether
the backpropagated signal $\partial \ell / \partial \hat{y}$, denoted by
$\texttt{backward} \in \reals^{\batchsize \times \numlabels}$ 
in the algorithm, is already zero, and if so skip the
corresponding operations. Note, in particular, that this means not only that the
multiplication with the zero value can be skipped, but it also makes it
unnecessary to load the second operand and to store the result. As sparse matrix
operations are highly memory-bound, this can be highly beneficial.

In fact, if we distribute the threads in the same way as the forward pass for
the calculation of the gradient with respect to the features (one thread
assigned for each \texttt{label} and \texttt{instance}) then most threads can be skipped
entirely.\footnote{On a GPU, skipping a single thread might not be helpful, as
threads are executed together in groups of 32 as a warp. However, with the very
high level of sparsity in the backward signal, it becomes common that all
threads within a warp can be skipped.} A schematic of the resulting
implementation is given in \autoref{alg:uniform-sparse-backward-features}.
Because multiple labels can contribute to the gradient of each input feature,
in this case several threads need to update the same part of the gradient array.
Therefore, we resort to using atomic addition operations here.

\begin{algorithm}
\begin{minted}{python}
out = backward[instance, label]
if out == 0:
    return

for weight_idx in range(s):
    source = indices[weight_idx, label]
    weight = weights[weight_idx, label]
    atomicAdd(gradient[instance, source], weight * out)
\end{minted}
\caption{Contribution to the gradient for the input features caused by a given
label and instance in the mini-batch.
\label{alg:uniform-sparse-backward-features}}
\end{algorithm}

For calculating the gradient of the weight values, it is possible to arrange
threads so that they can act independently, by using one thread for each
gradient entry, i.e., for each $\texttt{label} \in \intrange{\numlabels}$ and
$\texttt{weight\_idx} \in \intrange{s}$. In this case, one cannot skip entire
threads, but a zero in the backward signal still allows to skip the
unpredictable, indirect memory lookup of \texttt{feature = features
[instance, source]}, as shown in
\autoref{alg:uniform-sparse-backward-weights}.

\begin{algorithm}
\begin{minted}{python}
source = indices[weight_idx, label]
result = 0
for instance in range(batch_size):
    out = backward[instance, label]
    if out == 0: continue

    feature = features[instance, source]
    result += feature * out;
gradient[weight_idx, label] = result
\end{minted}
\caption{Calculation of the gradient for a given structural non-zero weight.
\label{alg:uniform-sparse-backward-weights}}
\end{algorithm}

\subsection{Mitigating underfitting by adding an intermediate layer}
Finally, we noticed that (even without uniformity constraint), replacing the
dense layer with a sparse layer results in diminished classification accuracy,
which we attribute to underfitting. Thus, we propose to improve the
expressiveness of the model by adding an intermediate layer between the
embedding layer and the final classification layer. Because the last layer is
sparse, its memory consumption is independent of the size of the preceding
layer. Thus, as long as this new intermediate layer is at least an order of
magnitude smaller than the number of labels, this does not impede our goal of
reducing memory requirements.

\section{Experiments}

\pgfplotstableset{
    highlightrow/.style={
        postproc cell content/.append code={
           \count0=\pgfplotstablerow
            \advance\count0 by1
            \ifnum\count0=#1
            \pgfkeysalso{@cell content/.add={\color{gray!80!black}}{}}
            \fi
        },
    },
}

\pgfplotstableset{simple-results-table/.style = {
    columns={setup,connectivity,intermediate,test-p@1,test-p@3,test-p@5,train-p@1,train-p@3,train-p@5,memory,epochs,time-per-epoch},
    every head row/.style={before row={\toprule \multicolumn{3}{c}{Setup}           & \multicolumn{3}{c}{Test} & \multicolumn{3}{c}{Train} & Mem. & Eps. & Time\\}, after row=\midrule,},
    every last row/.style={after row=\bottomrule}, 
    columns/setup/.style={string type, column name={Sparsity}, column type={l}},
    columns/connectivity/.style={column name={Con.}, column type={r}},
    columns/intermediate/.style={string type, column name={Int.}, column type={r@{\hskip 8pt}}},
    columns/train-p@1/.style={column name=P@1, fixed, fixed zerofill, precision=1, column type=r},
    columns/train-p@3/.style={column name=P@3, fixed, fixed zerofill, precision=1, column type=r},
    columns/train-p@5/.style={column name=P@5, fixed, fixed zerofill, precision=1, column type={r@{\hskip 8pt}}},
    columns/test-p@1/.style={column name=P@1, fixed, fixed zerofill, precision=1, column type=r},
    columns/test-p@3/.style={column name=P@3, fixed, fixed zerofill, precision=1, column type=r},
    columns/test-p@5/.style={column name=P@5, fixed, fixed zerofill, precision=1, column type={r@{\hskip 8pt}}},
    columns/memory/.style={column name={\si{\gibi\byte}}, fixed, fixed zerofill, precision=1, column type=r},
    columns/epochs/.style={column name={}, fixed, fixed zerofill, precision=1, column type=r},
    columns/time-per-epoch/.style={column name={sec}, fixed, fixed zerofill, precision=0, column type=r}
}}

\pgfplotstableset{size-results-table/.style = {
    simple-results-table,
    columns/train-p@1/.style={column name=P@1, string type, column type=r},
    columns/train-p@3/.style={column name=P@3, string type, column type=r},
    columns/train-p@5/.style={column name=P@5, string type, column type={r@{\hskip 8pt}}},
    columns/test-p@1/.style={column name=P@1, string type, column type=r},
    columns/test-p@3/.style={column name=P@3, string type, column type=r},
    columns/test-p@5/.style={column name=P@5, string type, column type={r@{\hskip 8pt}}},
    every row no  0/.style={before row={\multicolumn{12}{l}{\parbox[b][5mm][b]{0pt}{}\underline{\textsc{Slice Features}}}\\}},
    every row no  9/.style={before row={\multicolumn{12}{l}{\parbox[b][5mm][b]{0pt}{}\underline{\textsc{Cascade Features}}}\\}},
    highlightrow={1},
    highlightrow={10},
}}

In this section, we provide the experimental evidence showing that sparse last
layers are a viable approach to extreme multilabel classification. We run
experiments with several well-known benchmark datasets, measuring duration and
peak GPU memory consumption, as well as $\patk[k]$. After presenting results
that justify the architectural choices we made, we provide additional data
illustrating the trade-offs between memory consumption and classification
accuracy by varying the sparsity and size of the intermediate layer. Then we
present investigate the effect of implicit negative mining. The section
concludes with a discussion of the results.

\subsection{Experimental setup}
In this paper we focus on the setting of learning from fixed, low-dimensional
representations of the instances. This enables us to do many more experiments
than if we had to fine-tune an expensive transformer-based encoder for each run
of our model.

We use two different sources for the embeddings: 512-dimensional fast-text based
representations as used for \textsc{Slice}\cite{jain2019slice}, and the final
classification embeddings from a trained
\textsc{CascadeXML}\cite{kharbanda2022cascadexml} model with 768 dimensions. We
investigate on three datasets, \wikiten (\textsc{CascadeXML} only)
\cite{zubiaga_enhancing_2012}, \textsc{Amazon-670k} \cite{mcauley_hidden_2013},
and \wikifk \cite{Bhatia16}.

To update the network's weights, we use the \textsc{Adam}
optimizer\cite{kingma2014adam} with an initial learning rate of $\num{1e-3}$
that is decayed by $1/2$ whenever validation $\patk[3]$ stops improving, until
reaching $\num{1e-4}$. After that, training is stopped once $\patk[3]$ stops
increasing. For sparse layers, we initialize the connections uniformly randomly,
potentially subject to the constraint that each label gets the same amount of
connections. Every 1000 training steps\footnote{because
\wikiten has too few instances, we prune and redistribute at the beginning of each
epoch there}, each consisting of 32 samples in a minibatch, the
\SI{10}{\percent} lowest-magnitude weights are randomly redistributed. In order
to mitigate overfitting, we apply dropout to the input features, dropping
\SI{10}{\percent} for \amazonsix and \wikifkslice features, and
\SI{20}{\percent} for \wikiten and \wikifkcascade.

The experiments using with the larger datasets are run on a \textsc{Nvidia
V100}. Even though we want to demonstrate the feasibility of XMC learning on a
commodity GPU, in order to be able to make meaningful comparisons, we have to
train on the same GPU for all settings, which means that the GPU needs to have
enough memory to fit in a dense last layer. To quantify the memory benefits of
sparse training, we record the peak memory consumption as reported by tensorflow
(\verb|tf.config.experimental.get_memory_info("GPU:0")['peak']|). Note, in
particular, that all cases with our proposed architecture consume significantly
less than \SI{4}{\gibi \byte} of GPU memory, and thus will by feasible, albeit
training more slowly, on cheap gaming GPUs.

\subsection{Results with varying architecture}
\label{subsec:vary-arch}
As a first step, we want to show that the architectural choices described in 
\autoref{sec:method} are useful. To that end, we compare the training with a 
dense last layer to the following settings:
\begin{itemize}
    \item A single, unstructured sparse layer,
    \item A single, uniformly sparse layer,
    \item An intermediate, dense layer, followed by an unstructured sparse layer,
    \item An intermediate, dense layer, followed by a uniformly sparse layer.
\end{itemize}
The number of structural non-zeros is chosen such that in the \textsc{Unstructured}
sparse layers, there are an average of 32 connections per label, and in the
\textsc{uniform} sparse layers there are exactly 32 connections per label.

\begin{table}
\caption{Comparison of different network architectures. \texttt{Con} denotes the (average) number of connections per label,
\texttt{Int} the intermediate layer's size, \texttt{Mem} the peak GPU memory consumption, \texttt{Eps} the number of training 
epochs, and \texttt{Time} the duration of a single epoch in seconds. Bold marks the best results in any sparse setting. \textsc{Dense}
and \textsc{Uniform-32-32k} are averages of three runs for \amazonsix, the other entries are single realizations.}\label{results:very-arch}
\centering

\pgfplotstablevertcat{\datatbl}{data/Wiki10-cascade-arch.txt}
\pgfplotstablevertcat{\datatbl}{data/Wikipedia-500K-slice-arch.txt}
\pgfplotstablevertcat{\datatbl}{data/Wikipedia-500K-cascade-arch.txt}
\pgfplotstablevertcat{\datatbl}{data/Amazon670k-slice-arch.txt}
\pgfplotstablevertcat{\datatbl}{data/Amazon670k-cascade-arch.txt}
\pgfplotstabletypeset[simple-results-table,
every row no  0/.style={before row={\multicolumn{12}{l}{\parbox[b][5mm][b]{0pt}{}\underline{\wikitencascade}}\\}},
every row no  5/.style={before row={\multicolumn{12}{l}{\parbox[b][5mm][b]{0pt}{}\underline{\textsc{Wiki500k-Slice}}}\\}},
every row no  9/.style={before row={\multicolumn{12}{l}{\parbox[b][5mm][b]{0pt}{}\underline{\textsc{Wiki500k-Cascade}}}\\}},
every row no 12/.style={before row={\multicolumn{12}{l}{\parbox[b][5mm][b]{0pt}{}\underline{\textsc{Amazon670k-Slice}}}\\}},
every row no 17/.style={before row={\multicolumn{12}{l}{\parbox[b][5mm][b]{0pt}{}\underline{\textsc{Amazon670k-Cascade}}}\\}},
highlightrow={1},highlightrow={6},highlightrow={10},highlightrow={13},highlightrow={18},
    columns/train-p@1/.style={column name=P@1, string type, column type=r},
    columns/train-p@3/.style={column name=P@3, string type, column type=r},
    columns/train-p@5/.style={column name=P@5, string type, column type={r@{\hskip 8pt}}},
    columns/test-p@1/.style={column name=P@1, string type, column type=r},
    columns/test-p@3/.style={column name=P@3, string type, column type=r},
    columns/test-p@5/.style={column name=P@5, string type, column type={r@{\hskip 8pt}}},
]{\datatbl}
\end{table}

Unfortunately, due to the sheer size of the \wikifk dataset, and inefficient
training without intermediate layer (many epochs required) or without uniform
sparsity (very slow---up to \SI{4400} seconds/epoch), the training runs for
these setups timed out, and thus we do not have data for these settings.

The results of these experiments are presented in \autoref{results:very-arch}.
Several facts are immediately obvious from the recorded data: First, the naive,
tensorflow-based implementation for \textsc{Unstructured} sparsity is very slow,
to the degree that the sparse matrix multiplication ends up being $2$-$3\times$
slower than dense multiplication on the large datasets. Second, the
classification performance of uniform and unstructured sparsity is almost
identical (note that for \wikiten, the training stopped after the maximum of 200
epochs). Third, without an intermediate layer, there is a significant drop in
$\patk[k]$, both in training and test performance, showing that na\"ive
sparsification leads to severe overfitting. 

The measurements further show that for training based on \slice features, the
sparse implementation manages to attain and slightly surpass the classification
performance of the equivalent dense layer, whereas for \textsc{Cascade} features
there still remains a noticeable gap between dense and sparse training. As a
first possible explanation, one might argue that \textsc{Cascade} features have
been specifically trained so that they work well with a linear extreme
classification layer, whereas \slice are more general features. Therefore, it is
not the sparse realizations that perform better, but instead the dense setting
that performs disproportionately worse for \slice features, as it does not have
the benefit of the additional intermediate layer that allows non-linear
classification boundaries. This argument does not hold up, though, as both
features result in comparable model performance on the training set---it is the
\emph{generalization gap} that is much increased with \slice features.

Looking at the memory consumption, we can see that sparsification of the last
layer does lead to a noticeable reduction, but only becomes really effective
when we use our implementation of uniform sparsity. In this case, the memory
consumption reduces to between one third and on tenth of the dense equivalent.

\subsection{Results with varying network size}

In Tables~\ref{results:vary-size-amazon} and~\ref{results:vary-size-wiki}, we
demonstrate the effect of varying the number of connections per label, and the
size of the intermediate layer, for the uniformly sparse setup. Unsurprisingly,
increasing the network size results in improved classification performance. For
\textsc{Slice} features, the sparse network can be considerably better than the
dense counterpart. For \textsc{Cascade} features, increasing the size of the
sparse layer provides a way of shrinking the gap between sparse and dense
performance, while still remaining much more memory efficient than the dense
setup. In particular for \wikifk, the change in memory consumption is only by a
few percent, while the improvement in $\patk[k]$ is substantial. Except for
\amazonsix with \textsc{Cascade} features, increasing the model size results in
reducing the number of training epochs.

The data also shows a clear qualitative difference between \amazonsix and
\wikifk: For \amazonsix, switching from dense to sparse does not lead to a
noticeable decline in the ability of the classifier to fit the training set,
whereas for \wikifk the drop is dramatic, especially in the case of \slice
features. This suggests that for the smaller \amazonsix (\num{490449}
instances), even the sparse architectures are overparametrized enough to
interpolate the training set, whereas for \wikifk (\num{1813391} instances),
this is no longer the case, especially for the smaller sparse models.

\begin{table}
\caption{Train and test $\patk[k]$ on \amazonsix with varying sparsity and
intermediate-layer size, relative to dense performance.}\label{results:vary-size-amazon}
\centering
\pgfplotstablevertcat{\datatbl}{data/Amazon670k-slice-size.txt}
\pgfplotstablevertcat{\datatbl}{data/Amazon670k-cascade-size.txt}
\pgfplotstabletypeset[size-results-table]{\datatbl}
\end{table}

\begin{table}
\caption{Train and test $\patk[k]$ on \wikifk with varying sparsity and
intermediate-layer size, relative to dense performance.}\label{results:vary-size-wiki}
\centering
\pgfplotstablevertcat{\datatbl}{data/Wikipedia-500K-slice-size.txt}
\pgfplotstablevertcat{\datatbl}{data/Wikipedia-500K-cascade-size.txt}
\pgfplotstabletypeset[size-results-table]{\datatbl}
\end{table}

\subsection{Quantifying the effect of implicit negative mining}
Next, we show that the implicit negative mining effect discussed above can have
a significant impact on the speed of training. To that end, we use the small
model configuration with uniform sparsity with 32 structural non-zeros per
output and 16k intermediate units, and train it once using the squared hinge
loss (\textsc{Sqh}) and once using binary cross-entropy (\textsc{Bce}) loss
function. As the \textsc{Bce} loss only goes to zero asymptotically, this means
that there will not be many explicit zeros in the signal being back-propagated
through the sparse layer, and thus all labels have to be processed.

As shown in \autoref{results:bce-vs-sqh}, this has a strong effect on the
training time per epoch: The implicit negative mining with \textsc{Sqh} reduces
the duration by about one third. Additionally, the squared hinge loss results in
slightly better $\patk[k]$, and fewer training epochs. This finding is in
accordance with similar observations for extremely imbalanced classification
arising in object detection, where it was found that standard \textsc{Bce} is
outperformed by losses that give less weight to easy negatives, such as
\textsc{Focal loss}\cite{lin2017focal}.

\begin{table}
\caption{Comparison of training with square hinge loss and binary cross-entropy.}\label{results:bce-vs-sqh}
\centering
\pgfplotstablevertcat{\datatbl}{data/Wikipedia-500K-slice-bce.txt}
\pgfplotstablevertcat{\datatbl}{data/Wikipedia-500K-cascade-bce.txt}
\pgfplotstablevertcat{\datatbl}{data/Amazon670k-slice-bce.txt}
\pgfplotstablevertcat{\datatbl}{data/Amazon670k-cascade-bce.txt}
\pgfplotstabletypeset[simple-results-table,
columns={dataset,features,loss,test-p@1,test-p@3,test-p@5,train-p@1,train-p@3,train-p@5,memory,epochs,time-per-epoch},
columns/dataset/.style={string type,column type=l},
columns/features/.style={string type,column type=l},
columns/loss/.style={string type, column type={l@{\hskip 8pt}}},
]{\datatbl}
\end{table}

\subsection{Discussion}
The results above show that sparsification of the extreme layer is possible
without a strong decrease in classification performance, relative to a dense
layer. However, it has to be noted that training the dense layer in the common
experimental protocol employed here yields worse results than reported
state-of-the-art for the same set of features. Thus, even in cases where the 
sparse architecture outperforms the dense layer, reported results from the literature
are still better. 

In \autoref{tab:sota}, we present the results from \slice\cite{jain2019slice} and
\cascade\cite{kharbanda2022cascadexml}, compared against our largest setting
with 64 nonzeros per label and 65k intermediate units. Compared to these
methods, ours performs up to \SI{4}{\percent} worse, trading off a little
classification accuracy versus a multifold reduction in memory consumption. For
example, \cascade runs for over a day on two \textsc{NVidia A100} GPUs. 

\begin{table}
\caption{Comparison of sparse results with state-of-the-art.\label{tab:sota}}
\centering
\begin{tabular}{ll@{\hskip 10pt}ccc@{\hskip 10pt}ccc}
    \toprule
     &  & \multicolumn{3}{c}{\slice} & \multicolumn{3}{c}{\cascade} \\
    Dataset & Method & P@1 & P@3 & P@5 & P@1 & P@3 & P@5 \\
    \midrule
    \wikifk    & Literature & 62.6 & 41.8 & 31.6 & 77.0 & 58.3 & 45.1\\
    \wikifk    & Ours       & 60.5 & 39.8 & 29.8 & 74.5 & 56.0 & 43.2\\[4pt]
    \amazonsix & Literature & 37.8 & 33.8 & 30.7 & 48.8 & 43.8 & 40.1\\
    \amazonsix & Ours       & 34.6 & 30.5 & 27.7 & 45.3 & 39.8 & 35.9\\
    \bottomrule
\end{tabular}
\end{table}

\section{\textsc{Mach} as a special case of sparsity}

    The basic idea of \textsc{Mach} \cite{medini_extreme_2019} is to randomly group the $\numlabels$ labels into
    $\nummeta$ meta-labels, and then train an ensemble of meta-learners for
    different realizations of this grouping.%
    \footnote{The \textsc{Mach} paper works in the multi-class setting, which means that the following statements only hold 
    for if $\probability{\|\labelvec\|_1 = 1} = 1$. This can be achieved through a reduction of the original 
    multilabel problem \cite{menon_multilabel_2019}.
    }
    Let $\numhashings$ denote the
    number of these groups. The labels are mapped to meta-labels using
    2-universal hashing functions
    $\defmap{\hashfun^{(\hashindex)}}{\intrange{\numlabels}}{\intrange{\nummeta}}$.
    We can turn this into matrix form by using a family of matrices
    $\set{\indicatormat^{(\hashindex)} \in \set{0,1}^{\numlabels \times \nummeta}}_{\hashindex=1}^{\numhashings}$, such that
    \begin{equation}
        \indicatormatcmp^{(\hashindex)}_{\labelindex p} = 1 \,\Leftrightarrow\, \hashfun^{(\hashindex)}(\labelindex) = p \qquad \text{(label $\labelindex$ is in group $p$ for realization $\hashindex$)} \, .
    \end{equation}
    
    We thus can calculate the meta-label vectors as $\metalabelvec^{(\hashindex)} = \indicatormat^{(\hashindex)}{}^{\mathsf{T}} \labelvec$, which means for each component 
    $\metalabelvec_i^{(\hashindex)}(\labelvec) = \sum_{\labelindex=1}^{\numlabels} \indicatormatcmp^{(s)}_{\labelindex i} \labelvec_{\labelindex}$.
    The \textsc{Mach} algorithm then trains $\numhashings$ meta-classifiers $\defmap{\machcls^{(\hashindex)}}{\instancespace}{\closedinterval{0}{1}^{\nummeta}}$ that predict the probability, such that
    \begin{equation}
        \machcls^{(s)}_i(\rinstance) \approx \probability{g^{(s)}_i (\labelvec) = 1 \given \rinstance} \,.
    \end{equation}
    The key theorem in \cite{medini_extreme_2019} claims that, if this equation holds with exact equality, the original probabilities 
    can be recovered, as the following holds
    \begin{equation}
        \probability{\labelvec_{\labelindex} = 1 \given \rinstance} = \expectation{\frac{\nummeta}{\nummeta-1} \left( \frac{1}{\numhashings} \sum_{\hashindex=1}^{\numhashings} \machcls^{(\hashindex)}_{h^{(\hashindex)}(\labelindex)}(\rinstance) - \frac{1}{\nummeta} \right) \given \rinstance} \,,
        \label{eq:mach-thm}
    \end{equation}
    where the expectation is taken over the choice of randomized hash functions.

    We can rewrite the term $\machcls^{(\hashindex)}_{\hashfun^{(\hashindex)}(\labelindex)}(\sinstance)$ using the indicators $\indicatormat$ as 
    \begin{equation}
        \machcls^{(\hashindex)}_{h^{(\hashindex)}(\labelindex)}(\sinstance) = \sum_{t=1}^{\nummeta} c^{(\hashindex)}_{\labelindex t} \machcls^{(\hashindex)}_{t}(\sinstance) \,.
    \end{equation}

    This allows us to transform \eqref{eq:mach-thm} into a matrix equation: Let us consider the concatenated meta-predictions and concatenated (horizontally stacked) indicator,
    \begin{equation}
    \concatmachcls(\sinstance) \coloneqq \bigoplus_{\hashindex=1}^{\numhashings} \machcls^{(\hashindex)}(\sinstance) \in \reals^{\numhashings \cdot \nummeta} \,, \quad
    \indicatormat \coloneqq \bigoplus_{\hashindex=1}^{\numhashings} \indicatormat^{(\hashindex)} \in \reals^{\numlabels \times \numhashings \cdot \nummeta} \,,
    \end{equation}
    then we can combine these equations into
    \begin{equation}
        \expectation{\labelvec \given \rinstance} = \expectation{\frac{\nummeta}{\nummeta-1} \left( \frac{1}{\numhashings} (\indicatormat \cdot \concatmachcls(\sinstance)) - \frac{1}{\nummeta} \right)}.  \label{eq:mach-as-sparse}
    \end{equation}

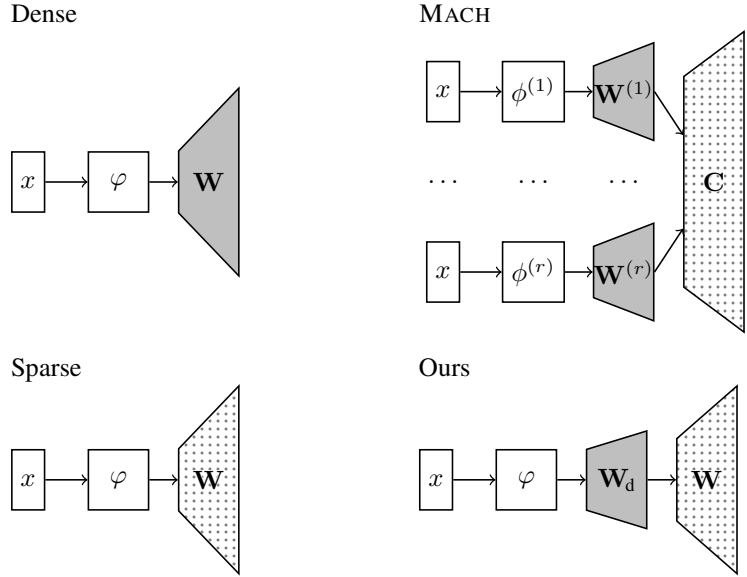
\begin{figure}
        \tikzset{input-node/.style={draw,rectangle, minimum height=0.8cm, minimum height=0.8cm}}
        \tikzset{weight-node/.style={draw,rectangle, minimum height=0.8cm, minimum width=0.8cm}}
        \tikzset{decode-node/.style={draw,trapezium, shape border rotate=90, minimum height=0.8cm, inner sep=0pt}}
        \tikzset{dense-style/.style={fill=gray!50!white}}
        \tikzset{sparse-style/.style={pattern=dots, pattern color=gray}}
        \tikzset{node distance=1.2cm}
        \centering
        \begin{tabular}{p{5cm}p{5cm}}
        Dense & \textsc{Mach} \\ 
        \begin{minipage}[c]{5cm}
        \begin{tikzpicture}[->, semithick]
            \node[input-node](x) {$x$};
            \node[weight-node, right of=x](w) {$\varphi$};
            \node[decode-node, dense-style, right of=w, minimum width=2.5cm, trapezium angle=60, trapezium stretches=true](out) {$\weightmatrix$};
            \draw (x) -- (w);
            \draw (w) -- (out);
        \end{tikzpicture}
        \end{minipage}
        \vspace{1ex}
        &
        \begin{minipage}[c]{5cm}
        \begin{tikzpicture}[->, semithick]
            \node[input-node](x) {$x$};
            \node[input-node, right of=x](p) {$\phi^{(1)}$};
            \node[below of=x](x2) {$\ldots$};
            \node[input-node, below of=x2](x3) {$x$};
            \node[decode-node, right of=p, minimum width=1.3cm, trapezium angle=80, trapezium stretches=true, dense-style](out1) {$\weightmatrix^{(1)}$};
            \node[right of=x2](mid) {$\ldots$};
            \node[right of=mid](mid2) {$\ldots$};
            \node[input-node, right of=x3](p3) {$\phi^{(r)}$};
            \node[decode-node, right of=p3, minimum width=1.3cm, trapezium angle=80, trapezium stretches=true, dense-style](out3) {$\weightmatrix^{(r)}$};
            \node[decode-node, sparse-style, right of=mid2, minimum width=4cm, trapezium angle=80, trapezium stretches=true](concat) {\contour{white}{$\indicatormat$}};
            \draw (x) -- (p);
            \draw (p) -- (out1);
            \draw (x3) -- (p3);
            \draw (p3) -- (out3);
            \draw (out.east) -- (concat);
            \draw (out3.east) -- (concat);
        \end{tikzpicture}
        \end{minipage}
        \vspace{1ex}
        \\
        Sparse & Ours \\ 
        \begin{minipage}[c]{5cm}
        \begin{tikzpicture}[->, semithick]
            \node[input-node](x) {$x$};
            \node[weight-node, right of=x](w) {$\varphi$};
            \node[decode-node, sparse-style, right of=w, minimum width=2.5cm, trapezium angle=60, trapezium stretches=true](out) {\contour{white}{$\weightmatrix$}};
            \draw (x) -- (w);
            \draw (w) -- (out);
        \end{tikzpicture}
        \end{minipage}
        &
        \begin{minipage}[c]{5cm}
        \begin{tikzpicture}[->, semithick]
            \node[input-node](x) {$x$};
            \node[weight-node, right of=x](w) {$\varphi$};
            \node[decode-node, right of=w, minimum width=1.3cm, trapezium angle=80, trapezium stretches=true, dense-style](w1) {$\weightmatrixd_{\!\text{d}}$};
            \node[decode-node, sparse-style, right of=w1, minimum width=2.5cm, trapezium angle=70, trapezium stretches=true](out) {\contour{white}{$\weightmatrix$}};
            \draw (x) -- (w);
            \draw (w) -- (w1);
            \draw (w1) -- (out);
        \end{tikzpicture}
        \end{minipage}
        \end{tabular}
        \caption{Dense training, \textsc{Mach} training, naive sparse architecture, and our proposed sparse architecture. Filled trapezoids indicate dense weight matrices, dotted indicates sparse
        weights. The matrix $\indicatormat$ in \textsc{Mach} is fixed, the different $\weightmatrix$ matrices are learnable.}
        \label{fig:models}
    \end{figure}

    That means the we can \emph{interpret} the inference procedure of MACH as
    performing a large, but very sparse, matrix multiplication. In MACH, the
    matrix $\indicatormat$ is fixed, with a certain block structure, and the
    corresponding blocks of coordinates in $\concatmachcls$ are trained
    \emph{independently}. This has the advantage that the algorithm is trivially
    paralellizable, but it also imposes three important restrictions: First, the
    different $\machcls$'s cannot adapt to each other due to the lack of end-to-end
    training. Second, since this approach requires $\numhashings$ copies of
    $\machcls$, this network cannot be too large itself. Third, as \eqref{eq:mach-thm}
    is the expectation over randomized hash functions, the equality only holds in 
    the limit of infinite number of meta-classifiers. 
    A schematic comparison of our architecture to (i) a dense last layer, 
(ii) a vanilla sparse last layer, and (iii) \textsc{Mach} (as an inference strategy) is shown in \autoref{fig:models}.

\section{Conclusion and Outlook}

In this paper, we have showed that it is possible to replace an extreme-scale
dense classification layer with a memory-efficient sequence of an
intermediately-sized layer followed by a uniformly-sparsely connected layer, 
without a strong drop in classification performance, and in some cases even 
improved $\patk[k]$.

The experiments performed so far investigate sparse layers in the context of a
simple training procedure: Learning with the full label space, from fixed,
pre-trained features. To achieve feature-parity with existing approaches, this
needs to be extended to allow for end-to-end training, where the featurizer
$\featureextract$ is learned jointly with the classifier. Secondly, even though
the implicit negative mining effect allows to reduce the computation for the
backward pass to be sub-linear in the overall number of labels, it still
requires a full forward pass. In order to get to competitive training times, one
thus has to integrate also explicit negative mining into the training pipeline.

We believe that this paper provides a good foundation, from which these goals
can be achieved: First, by having the sparse multiplication implemented as a
regular tensorflow layer, it can be readily included in a more general model,
and automatic differentiation will ensure correct gradient calculations. Second,
because we are constraining the sparsity to be uniform, selecting a subset of
labels for which scores shall be calculated becomes a trivial matrix slicing
operation, similar to the fully-connected case. 
Furthermore, from a statistical perspective, it is possible that uniform sparsity also leads to a better coverage of tail-labels, and improvements in the corresponding metrics \cite{jain2016extreme, schultheis2022missing}.
In the followup works, one could incorporate the proposed framework into existing end-to-end deep extreme classification frameworks while benefiting from explicit negative mining.

\section{Acknowledgements}
We acknowledge the support of computational resources provided by the Aalto Science-IT project, and CSC IT Center for Science, Finland. This work is funded in part by the Academy of Finland projects : 347707 and 348215.
%
%
%
\bibliographystyle{splncs04}
\bibliography{lit}

\end{document}